\documentclass[runningheads]{llncs}

\usepackage[T1]{ fontenc }
\usepackage[dvipsnames]{ xcolor }
\usepackage[normalem]{ ulem }
\usepackage[pdftex]{ graphicx }
\usepackage[inline]{ enumitem }
\usepackage{ booktabs }
\usepackage{ siunitx }
\usepackage{ optidef }
\usepackage{ multirow }
\usepackage[english]{ babel }
\usepackage{ algorithm, algorithmic }
\usepackage{ makecell }
\usepackage{ url }
\usepackage{ wrapfig2 }
\usepackage{ adjustbox }
\usepackage{ ifthen }
\usepackage{ amsmath, amssymb, dsfont, mathtools, amsfonts, nicefrac, microtype, bm }
\usepackage{ xspace }
\usepackage[makeroom]{cancel}

\usepackage[misc]{ifsym}

\usepackage[numbers]{natbib}

\usepackage[acronym]{ glossaries }
\glsdisablehyper

\newacronym{smns}{SNs}{Social Networks}
\newacronym{smn}{SN}{Social Network}

\newacronym{ios}{IOs}{Information Operations}
\newacronym{io}{IO}{Information Operation}

\newacronym{tpp}{TPP}{Temporal Point Process}
\newacronym{tpps}{TPPs}{Temporal Point Processes}

\newacronym{llm}{LLM}{Large Language Model}
\newacronym{llms}{LLMs}{Large Language Models}

\newacronym{model}{TENSOR}{Temporal-bEhavior-laNguage Signals for information Operation Recognition}

\newacronym{ad}{AD}{Anomaly Detection}

\newacronym{ephad}{EPHAD}{Evidence-based Post-Hoc Adjustment Framework for Anomaly Detection}

\newacronym{nll}{NLL}{negative log-likelihood}

\newacronym{sahp}{SAHP}{Self-attentive Hawkes Process}
\newacronym{gpt}{gpt-oss-120B}{gpt-oss-120B}
\newacronym{llama}{llama3.3-70B}{llama3.3-70B}
\newacronym{qwen}{qwen-next-80B}{qwen-next-80B}
\newacronym{glm}{glm-4.5-air}{glm-4.5-air}
\newacronym{mistral}{mistral3.2-24b}{mistral3.2-24b}

\newacronym{luceri}{LLM}{LLM-based \acrshort{io} user detection}
\newacronym{kong}{Clustering}{User Embedding Clustering}
\newacronym{nwala}{BLOC}{Behavioral Languages for Online Characterization}

\newacronym{auc}{AUC}{Area Under the receiver operating characteristic Curve}
\newacronym{ap}{AUPRC}{Area Under the Precision-Recall Curve}

\usepackage{ hyperref }
\usepackage[capitalise, noabbrev]{ cleveref }

\glsdisablehyper  
\newcommand{\partopic}[1]{\paragraph{#1}}

\NewCommandCopy{\oldcitep}{\citep}
\renewcommand{\citep}[1]{~\oldcitep{#1}}

\newcommand{\meanstd}[2]{#1\(\pm\)#2}

\newcommand{\mysection}[2][]{
    \ifthenelse{ \equal{#1}{} }
    {\section{\texorpdfstring{#2}{#2}}}
    {\section{\texorpdfstring{#2}{#1}}}
}
\newcommand{\mysubsection}[2][]{
    \ifthenelse{ \equal{#1}{} }
    {\subsection{\texorpdfstring{#2}{#2}}}
    {\subsection{\texorpdfstring{#2}{#1}}}
}
\newcommand{\mysubsubsection}[2][]{
    \ifthenelse{ \equal{#1}{} }
    {\subsubsection{\texorpdfstring{#2}{#2}}}
    {\subsubsection{\texorpdfstring{#2}{#1}}}
}

\newcommand{\corr}{(\Letter)}
 
\makeatletter
\DeclareRobustCommand\onedot{\futurelet\@let@token\@onedot}
\def\@onedot{\ifx\@let@token.\else.\null\fi\xspace}

\def\ie{\emph{i.e}\onedot, }

\makeatother

\newcommand{\data}[1]{\bm{#1}}

\DeclareMathAlphabet{\mathsfit}{\encodingdefault}{\sfdefault}{m}{sl}
\SetMathAlphabet{\mathsfit}{bold}{\encodingdefault}{\sfdefault}{bx}{n}

\newcommand{\expect}{\mathbb{E}}

\NewCommandCopy{\oldexp}{\exp}
\renewcommand{\exp}[1]{\oldexp\left(#1\right)}

\newcommand{\model}{\mathcal{M}}
\newcommand{\history}{\bm{\mathcal{H}}}
  
\begin{document}
\title{Unsupervised Anomaly Detection of Information Operations Users via Behavioral and Language Patterns}
\toctitle{Unsupervised Anomaly Detection of Information Operations Users via Behavioral and Language Patterns}

\titlerunning{IO User Detection via Behavior and Language Patterns}

\author{Sishun Liu\inst{1}\and
Sajal Halder\inst{1} \and
Ke Deng\inst{1} \and Yan Wang\inst{2} \and Xiuzhen Zhang\inst{1} \corr}
\tocauthor{Sishun Liu, Sajal Halder, Ke Deng, Yan Wang, Xiuzhen Zhang}

\authorrunning{S. Liu et al.}

\institute{RMIT University, Melbourne, Victoria 3000, Australia \email{sishun.liu@student.rmit.edu.au, \{sajal.halder,ke.deng,xiuzhen.zhang\}@rmit.edu.au}
\and
Macquarie University, Sydney, New South Wales 2000, Australia
\email{yan.wang@mq.edu.au}}

\maketitle   
 
\begin{abstract}

\textit{\acrfull{ios}} on \textit{\acrfull{smns}} have been identified as a significant threat to democracy and modern society, but they are challenging and expensive to detect by humans. Existing supervised \acrshort{io} detection methods fail to capture the dynamic nature of evolving \acrshort{io} user behavior, while existing unsupervised approaches rely on oversimplified assumptions of coordination among \acrshort{io} users that may not exist in practice. To overcome the limitations of existing methods, we formulate \acrshort{io} user detection as an anomaly detection problem and propose a novel unsupervised \acrshort{io} user detection approach called \textit{\acrfull{model}}, which leverages multimodal data, including temporal online user behavior, such as message posting activities, and the textual content of the messages. The motivation is that \acrshort{io} users are typically a very small fraction of all online users and have unique temporal behavioral and language patterns. Specifically, we train a \textit{\acrfull{tpp}} to capture abnormal temporal behavioral patterns of \acrshort{io} users because they are known to behave in a coordinated manner for \acrshort{io} campaigns. We further introduce a novel \textit{evidence function} that converts \acrshort{llm} responses, which are generated from user post timelines, into quantitative scores to adjust the \acrshort{tpp} outputs for better \acrshort{io} user detection. Experimental results show that \acrshort{model} outperforms the baselines on five real-world \acrshort{io} datasets\footnote{Code is available at \url{https://github.com/xiuzhenzhang/TENSOR}.}.

\keywords{\acrlong{io} User Detection \and \acrlong{tpp} \and \acrlong{llm}}

\end{abstract}
 
\mysection{Introduction}
\label{sec:intro}

The development of \textit{\acrfull{smns}} enables fast dissemination of critical information, large-scale discussions, and joint actions about political and social issues because \acrshort{smns} connect people\citep{ezzeddineExposingInfluenceCampaigns2023}.
However, the capabilities of \acrshort{smns} can be misused by \textit{\acrfull{ios}}, especially state-sponsored ones.
\acrshort{ios} are organized attempts to tamper with the regular flow of information and influence public opinion through disinformation, hate speech, and other harmful content.
\acrshort{ios} are hard to detect because they are always initiated by a small group of users\citep{seckinLabeledDatasetsResearch2025a, vishnuprasadTrackingFringeCoordinated2024}.
With targets including narrative manipulation and the fostering of division in online and real-world communities, IOs have been identified as a significant threat to democracy, and the need for robust methods to detect these operations is urgent\citep{temporary-citekey-5930}.

Researchers have proposed \acrshort{io} detection approaches to identify whether individual posts or specific users are related to an \acrshort{io}. The following discussion focuses on the detection of \acrshort{io} users, which is the primary interest of this study.
\textit{\acrshort{io} users} are motivated or incentivized to promote \acrshort{ios}, while legitimate organic users are called \textit{control users}.
\acrshort{io} user detection approaches use patterns within user post timelines on \acrshort{smns}.
These patterns can be categorized as \textit{behavioral patterns} (specifically, \textit{temporal} behavioral patterns because they describe the user activities on \acrshort{smns} over time) and \textit{language patterns}, including speaking style and areas of interest.

\acrshort{io} user detection is challenging.
The biggest challenge is the generalization capability of \acrshort{io} user detection algorithms, \ie their ability to detect unseen \acrshort{ios}.
In the wild, \acrshort{ios} evolve quickly, so existing labeled IO datasets always lag.
This means that existing supervised\citep{addawoodLinguisticCuesDeception2019a, ezzeddineExposingInfluenceCampaigns2023, luceriDetectingTrollBehavior2020} and semi-supervised\citep{alizadehContentbasedFeaturesPredict2020, miniciIOHunterGraphFoundation2025, smithAutomaticDetectionInfluential2021, vargasDetectionDisinformationCampaign2020} \acrshort{io} user detection methods suffer from a generalization issue, \ie they cannot detect new, unseen \acrshort{ios}. Recent zero-shot \acrshort{llm} approaches\citep{luceriLeveragingLargeLanguage2024} provide an unsupervised alternative, but their ability to capture complex temporal behavioral dynamics is still limited.

This paper formulates \acrshort{io} user detection as an unsupervised anomaly detection problem and proposes a novel approach, called \textit{\acrfull{model}}, to use behavioral and language patterns from user post timelines for unsupervised \acrshort{io} user detection.
Training anomaly detection models on \acrshort{io} data is difficult because the training data, which are supposed to consist of timelines of control users, are contaminated by \acrshort{io} users. 
This harms the performance of \acrshort{io} user detection models.
To mitigate this issue, we use language patterns.
Specifically, first, we train a \textit{\acrfull{tpp}} on the contaminated data to identify \acrshort{io} users based on abnormal behavioral patterns, such as coordinated behavior among different \acrshort{io} accounts, which are commonly observed among \acrshort{io} users but rare among control users.
A \acrshort{tpp} is a well-defined stochastic process over temporal event sequences\citep{daley_introduction_2003}.
Researchers have used the process to achieve state-of-the-art performance in unsupervised identification of abnormal event sequences from normal ones\citep{shchurDetectingAnomalousEvent2021} and outlier events from normal events\citep{liuEventOutlierDetection2021}.
Second, we propose a novel \textit{evidence function} to adjust \acrshort{tpp} inference. This function converts \acrshort{llm} responses, which are generated from user post timelines, into quantitative scores that refine the \acrshort{tpp} output for better \acrshort{io} user detection.
Experimental results show that \acrshort{model} outperforms other baselines by a significant margin on five real-world \acrshort{io} datasets.
 
\mysection{Related Work}
\label{sec:works}

Most existing \acrshort{io} detection studies classify \acrshort{io} users based on pure behavioral patterns\citep{ezzeddineExposingInfluenceCampaigns2023, luceriDetectingTrollBehavior2020}, pure language patterns\citep{addawoodLinguisticCuesDeception2019a, alizadehContentbasedFeaturesPredict2020, haiderDetectingSocialMedia2023, imStillOutThere2020, jachimTrollHunterEvaderAutomated2020}, or both\citep{luceriLeveragingLargeLanguage2024, miniciIOHunterGraphFoundation2025, vargasDetectionDisinformationCampaign2020}.
Vargas et al.\citep{vargasDetectionDisinformationCampaign2020} classify \acrshort{io} users based on coordination behaviors, such as tweeting the same content within a timeframe, or tweeting the same hashtag within a timeframe, extracted from users' post timelines.
Luceri et al.\citep{luceriLeveragingLargeLanguage2024} investigate \acrshort{io} user detection for \acrshort{llms} under few-shot and fine-tuning settings. 
Minici et al.\citep{miniciIOHunterGraphFoundation2025} propose IOHunter, an \acrshort{io} detector built upon a graph, in which the similarity of users' behavioral traces, including tweets, hashtags, and time of posting, justifies connections between users. Despite reported good performance on existing data, supervised and semi-supervised \acrshort{io} user detection methods require labelled \acrshort{io} datasets, which are limited in size and lag behind real-world \acrshort{io} practices. This raises concerns about the generalization capabilities of supervised and semi-supervised \acrshort{io} user-detection algorithms in real-world scenarios. 

Recent zero-shot \acrshort{llm} approaches\citep{luceriLeveragingLargeLanguage2024} provide an unsupervised alternative, but their ability to capture complex temporal behavioral dynamics is still limited. Although not directly aimed at \acrshort{io} user detection, other studies concerning \acrshort{io} users verify coordination among users in the same \acrshort{io}\citep{kongIntervalcensoredTransformerHawkes2023, nwalaLanguageFrameworkModeling2023}. Although such methods reveal that \acrshort{io} users tend to form small and relatively well-separated clusters compared with legitimate organic users, our study shows that it is an oversimplification to develop unsupervised \acrshort{io} user detection methods solely based on this observation.

\textit{Anomaly Detection using contaminated data:} Anomaly detection using contaminated data is possible if the anomalies occupy a small portion of the data\citep{wang_effective_2019}.
Most existing methods employ unsupervised identification of anomalies during training, allowing the model to either exclude these samples or leverage them to mitigate performance degradation\citep{perini_estimating_2023, qiuLatentOutlierExposure2022, yoon_self-supervise_2022}.
Patra et al.\citep{patraEvidenceBasedPostHocAdjustment2025} observed that these methods require prior knowledge about the dataset, such as the contamination ratio, which is usually unknown.
To solve these issues, they proposed \textit{\acrfull{ephad}}, the first framework to adjust the prediction of an anomaly detector trained on contaminated data using an evidence function during test time.
Patra et al.\citep{patraEvidenceBasedPostHocAdjustment2025} focus on anomaly detection on visual data, but how to adapt \acrshort{ephad} to multimodal temporal and language data remains an open question.

\textit{\acrshort{llms} for Social Media Analysis:} \acrshort{llm}-based social media analysis has been widely investigated, including post annotation\citep{malik_pseudo-labeling_2024, tekumalla_leveraging_2023, tornberg_large_2025}, misinformation detection and mitigation\citep{kumar_silver_2025, qi_sniffer_2024, zhang_toward_2024, zhou_correcting_2026}, and \acrshort{io} detection\citep{luceriLeveragingLargeLanguage2024}.
Unlike \citep{luceriLeveragingLargeLanguage2024}, which uses \acrshort{llms} as the sole classifier based on behavioral and language patterns, \acrshort{model} leverages an \acrshort{llm} to provide additional evidence to enhance the \acrshort{tpp}-based anomaly detector. 
\mysection{Problem Definition}
\label{sec:pro_def}

We formulate \acrfull{io} user detection as an unsupervised anomaly detection problem.
A \acrfull{smn} contains \acrshort{io} users, \ie users motivated or incentivized to promote an \acrshort{io}, and control users, \ie normal genuine social media users.
The number of \acrshort{io} users is significantly smaller than that of normal users, and their behavioral and language patterns are noticeably different, as reflected in their \textit{post timeline}. 
The post timeline of user \(i\) is \(\data{s}_i = (s_{i,1}, s_{i,2}, \cdots, s_{i,M_i})\), where \(M_i\) is the number of posts by this user.
Each post event \(s_{i,k} = (t_{i,k}, c_{i,k})\) consists of a time \(t_{i,k}\) when this post is created and post content \(c_{i,k}\).
Post timelines of all users, including normal and \acrshort{io} users, are denoted as \(\data{S}=\{\data{s}_1, \data{s}_2, \cdots, \data{s}_n\}\).
For later use, we denote the respective sequences of all timestamps and contents as \(\data{T} = \{\data{t}_1, \data{t}_2, \dots, \data{t}_n\}\) and \(\data{C} = \{\data{c}_1, \data{c}_2, \dots, \data{c}_n\}\), where \(\data{t}_i = (t_{i,1}, t_{i,2}, \cdots, t_{i,M_i})\) and \(\data{c}_i = (c_{i,1}, c_{i,2}, \cdots, c_{i,M_i})\) are sequences of timestamps and contents of user \(i\).
We train a model \(\model\) on \(\data{S}\) to identify \acrshort{io} users from control users.
\acrshort{io} users are labeled 0, and control users are labeled 1, formulated as follows:
\begin{equation}
y = 
  \begin{cases}
    0, & \model(\data{t}_i, \data{c}_i) \geqslant\epsilon \\
    1, & \text{otherwise}
  \end{cases}
\label{eqn:definition}
\end{equation}
where \(\epsilon\) is the threshold. We show how we determine the value of \(\epsilon\) in \cref{sec:exp}.
 
\mysection[Methodology]{Methodology}
\label{sec:methodology}
In this section, we introduce \acrfull{model}, a novel anomaly detection approach that uses behavioral and language patterns from user action sequences for unsupervised \acrshort{io} user detection.
\acrshort{model} consists of two core modules: \acrfull{tpp} and \acrfull{llm}.
\acrshort{tpp} identifies \acrshort{io} users using user action sequences, but its performance can be suboptimal because the training set is contaminated by \acrshort{io} users.
To mitigate this issue, we propose a novel evidence function to convert \acrfull{llm} outputs into a score and use \textit{\acrfull{ephad}}\citep{patraEvidenceBasedPostHocAdjustment2025} to adjust the output of the \acrshort{tpp} model using the evidence function.
In \cref{sec:train_tpp} and \cref{sec:evidence_func}, we provide brief introductions to \acrshort{tpp} and \acrshort{ephad}.
In \cref{sec:model}, we present the structure and technical details of \acrshort{model}.

\mysubsection[TPP model]{\acrshort{tpp} model}
\label{sec:train_tpp}
The \acrfull{tpp} describes a random process of an event sequence \(\boldsymbol{\tau} = (\tau_1, \tau_2, \cdots, \tau_m)\) where \(\tau_k\) is the occurrence time.
This paper considers the simple \acrshort{tpp}, which only allows at most one event at any time, thus \(\tau_k<\tau_\ell\) if \(k<\ell\).
Given the history up to (exclusive) the current time \(t\), denoted as \(\history\), the \textit{conditional intensity function} \(\lambda^*(t)\) is the probability that an event will happen at time \(t\)\citep{daley_introduction_2003}:\footnote{The asterisk denotes that this function conditions on history.}
\begin{equation}
\label{eqn:def_tpp_intensity}
    \lambda^*\left(t\right) = \lim_{\Delta t \rightarrow 0}{\dfrac{P\left(\tau \in (t, t+\Delta t]\middle|\history\right)}{\Delta t}}
\end{equation}
With \(\lambda^*(t)\), we can define the joint probability distribution \(p^*(t)\) of the next event at \(t\).
\begin{equation}
    \label{eqn:tpp_p}
    p^*\left(t\right) = \lambda^*\left(t\right)\exp{-\int_{t_l}^{t}{\lambda^*(\tau)\mathrm{d}\tau}}
\end{equation}
The \acrfull{nll} loss on \(\boldsymbol{\tau}\) observed in a time interval \([t_0,T]\) is:
\begin{equation}
\label{eqn:nll_of_tpp}
    L = -\log p(\boldsymbol{\tau}) = - \sum_{k = 1}^{M}{\log \lambda^*(\tau_k)} + \int_{t_0}^{T}{\lambda^*(u)\mathrm{d}u}
\end{equation}
where \(M\) is the number of events in \(\boldsymbol{\tau}\). \cref{eqn:nll_of_tpp} is the training loss of \acrshort{tpp} models. Recent \acrshort{tpp} models are based on neural networks (see \citep{shchurNeuralTemporalPoint2021} for a comprehensive survey).
\acrshort{tpp} has been used to detect abnormal events or sequences from normal events or event sequences\citep{liuEventOutlierDetection2021, shchurDetectingAnomalousEvent2021, zhangMultipleHypothesisTesting2023}.

\mysubsection[EPHAD]{\acrlong{ephad}}
\label{sec:evidence_func}
\acrfull{ad} algorithms need a training dataset \(\data{D} = \{x_1, x_2, \cdots, x_n\}\) containing normal samples to train an \acrshort{ad} classifier \(f(x)\).
If \(\data{D}\) is contaminated by abnormal data, the trained \acrshort{ad} model \(f(x)\) will treat anomalies as normal.
Patra et al.\citep{patraEvidenceBasedPostHocAdjustment2025} observed that \acrshort{ad} models trained on contaminated \(\data{D}\) can be corrected at test time using a predefined evidence function \(T(x)\).
The \(T(x)\) contains external knowledge about anomalies, which is not captured by \(f(x)\).
The method is called \textit{\acrfull{ephad}}.
Specifically, \acrshort{ephad} revises \(f(x)\) using \textit{exponential tilting} as:
\begin{equation}
    \hat{f}(x) = \frac{f(x)\exp{T(x)/\beta}}{Z}
    \label{eqn:ephad}
\end{equation}
where \(\beta > 0\) is the temperature, and \(Z=\int_{D}{f(x)\exp{T(x)/\beta}dx}\) is the normalizing constant.
This method is theoretically grounded in Korbak et al.\citep{korbakRLKLPenalties2022}.
They treat the Post-Hoc adjustment as a KL-regularized optimization problem, where the objective is to find an adjusted distribution \(\hat{f}(x)\) that minimizes the KL divergence from the original distribution \(f(x)\) while maximizes the expected evidence \(T(x)\), as shown in the following optimization problem:
\begin{equation}
    \max_{\hat{f}} \expect_{x\sim\hat{f}}[T(x)] - \beta D_{\textrm{KL}}(\hat{f} \| f)
\end{equation}
where \(D_{\textrm{KL}}(\hat{f} \| f)\) is the KL-divergence between \(\hat{f}\) and \(f\).
They show that \cref{eqn:ephad} is the solution to this optimization problem.
Because \acrshort{ad} only depends on the relative ordering of samples, the intractable normalizing constant \(Z\) in \cref{eqn:ephad} can be omitted, which simplifies \cref{eqn:ephad} into:
\begin{equation}
    \hat{f}(x) = f(x)\exp{T(x)/\beta}
    \label{eqn:ephad_sim}
\end{equation}
This is more practical for implementing \acrshort{ad} models. In our approach elaborated in the following section, we use \cref{eqn:ephad_sim} to adjust \(f(x)\) based on \acrshort{tpp} models trained on contaminated training data using the evidence function derived from \acrshort{llm} outputs for better \acrshort{io} user detection.

\subsection[Our Model]{\acrlong{model}}
\label{sec:model}

\begin{figure}[!t]
    \centering
    \includegraphics[width=0.95\textwidth]{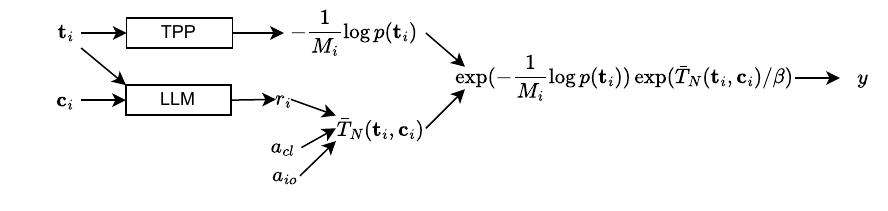}
    \caption{Architecture of \acrfull{model}.}
    \label{fig:model}
\end{figure}
 
In this section, we propose \acrfull{model}, which is sketched in \cref{fig:model}.
\acrshort{model} checks whether a user is an \acrshort{io} user based on their \(\data{t}_i\) and \(\data{c}_i\).
The abstract input \(x\) in \cref{eqn:ephad_sim} is instantiated as a user \(x_i=(\data{t}_i, \data{c}_i)\), where \(\data{t}_i\) and \(\data{c}_i\) represent the sequence of timestamps and contents of user \(i\), respectively.
It has two main components.
The first component is a frozen \acrshort{tpp} model trained on \(\data{T}\), which contains both \acrshort{io} and control users.
The \acrshort{tpp} model captures the difference between control and \acrshort{io} users using behavior patterns, including the coordinated behavior between \acrshort{io} accounts or an excessive number of posts compared with control users.
The differences are reflected in the value \(-\frac{1}{M_i} \log p(\data{t}_i)\).
Thus, \(f(x_i)=\exp{-\frac{1}{M_i}\log p(\data{t}_i)}\) is the \acrshort{tpp}-based score computed from the timestamp component of \(x_i\).
The second component is the averaged evidence function \(\bar{T}_N(\data{t}_i, \data{c}_i)\), backed by a frozen \acrshort{llm}.
It determines whether a user is an \acrshort{io} user based on behavioral and language patterns in \(\data{c}_i\) and \(\data{t}_i\).
That is, \(T(x_i)=\bar{T}_N(\data{t}_i, \data{c}_i)\) is the averaged evidence function over \(N\) samples derived from a frozen \acrshort{llm}.
\(-\frac{1}{M_i} \log p(\data{t}_i)\) and \(\bar{T}_N(\data{t}_i, \data{c}_i)\) are later fused using \acrshort{ephad} to mitigate the effect of \acrshort{tpp} trained on contaminated \(\data{T}\).
The outputs of the two components are fused using \cref{eqn:ephad_sim} to mitigate the effect of data contamination in \acrshort{tpp}.

\acrshort{model} uses \acrshort{tpp} models to classify \acrshort{io} users using the behavioral patterns collected from \(\data{t}_i\).
Because there are more control users than \acrshort{io} users, \acrshort{tpp} models yield higher probabilities \(p(\data{t}_i)\) for control users compared to \acrshort{io} users.
This makes \acrshort{tpp} a valid \acrshort{ad} classifier:
\begin{equation}
y = 
  \begin{cases}
    0, & \text{if}\ -\frac{1}{M_i}\log p(\data{t}_i) \geqslant \epsilon \\
    1, & \text{otherwise}
  \end{cases}
\label{eqn:ad_mtpp}
\end{equation}
where \acrshort{io} users are labeled \(0\), control users are labeled \(1\), \(\epsilon\) is the threshold, and \(M_i\) is the length of \(\data{t}_i\).

\begin{figure}[!t]
    \centering
    
    \noindent\centering
    \fbox{
        \begin{minipage}{\linewidth}
            \setlength{\parskip}{0.2em}
            You are an advanced social media analyst specialized in detecting Information Operations (IO). Your objective is to analyze user profiles and posting behaviors to distinguish between state-sponsored Information Operation (IO) accounts and control users based on the following behavioral frameworks:
            
            Role A: Information Operation (IO) Account

            Identity: These accounts are verified as part of inauthentic, coordinated efforts backed by state actors to manipulate public debate.
            
            Tactics:
            
            * Strategic Manipulation: They employ tactics like hashtag hijacking, artificial amplification, and the dissemination of propaganda or disinformation.
            
            * Targeting: They focus on specific audience communities, often using coordinated actions such as flooding through political cartoons or memes.
            
            * Profile Composition: They may consist of human-operated accounts, automated bots, or compromised profiles that have been repurposed for a campaign.
            
            Role B: Control Account

            Identity: These represent legitimate, organic users who act as a baseline for "normal" social media behavior.
            
            Selection Context: These users are identified by their engagement in the same topics and hashtags as IO accounts during the same time frames, but without coordination.
            
            Behavioral Characteristics:

            * Authentic Engagement: They discuss similar political or social topics without endorsing or participating in an orchestrated state agenda.
            
            * Content Diversity: Their timelines include posts on unrelated personal or general topics, whereas IO accounts are often more single-mindedly focused on campaign goals.
            
            IO accounts are rare. Most accounts are control accounts.
            
            Please do your analysis step by step. If you can think, think as thoroughly as possible to make your decision since the results are quite important. Your thinking process can be long, but the final response should be concise: only answer whether this account is an IO account or a control account.
        \end{minipage}
    }
    \caption{Prompts for zero-shot \acrshort{io} user detection}
    \label{fig:prompt}
\end{figure}
 
However, the dataset \(\data{S}\) is contaminated by \acrshort{io} users, which could harm detection accuracy.
To mitigate this, \acrshort{model} adjusts \(-\frac{1}{M_i}\log p(\data{t}_i)\) using a \acrshort{llm}-based evidence function \(T(\data{t}_i, \data{c}_i)\).
The \acrshort{llm} takes in the system prompt in \cref{fig:prompt} and a structured input of \(\data{t}_i\) and \(\data{c}_i\).
It then generates a response \(r_i = \mathrm{LLM}(\data{t}_i, \data{c}_i)\), indicating whether the user is an \acrshort{io} user or a control user by answering ``\acrshort{io} account'' or ``Control account''.
To convert \(r_i\) into \(T(\data{t}_i, \data{c}_i)\), we use a \textit{semantic similarity mapping} strategy instead of enforcing strict text matching, which is prone to failure because \acrshort{llm} outputs are non-deterministic.
Specifically, we compute the semantic similarity between the raw \acrshort{llm} output \(r_i\) and reference texts representing \acrshort{io} or control users.
In our case, the reference texts are ``\acrshort{io} account'' for \acrshort{io} users and ``Control account'' for control users, denoted as \(a_{io}\) and \(a_{cl}\).
By applying a \(\mathtt{softmax}\) transformation to these similarity scores, we obtain a probability distribution over the \acrshort{io} and control users.
\(T(\data{t}_i, \data{c}_i)\) is defined as the probability of selecting the \acrshort{io} user:
\begin{equation}
  \label{eqn:evidence_func}
  T(\data{t}_i, \data{c}_i) = \frac{\exp{\mathrm{sim}(a_{io}, \mathrm{LLM}(\data{t}_i, \data{c}_i))}}{\exp{\mathrm{sim}(a_{cl}, \mathrm{LLM}(\data{t}_i, \data{c}_i))} + \exp{\mathrm{sim}(a_{io}, \mathrm{LLM}(\data{t}_i, \data{c}_i))}}
\end{equation}
where \(\mathrm{sim}(x, y)\) measures the similarity between input text \(x\) and \(y\).
The output of \acrshort{llms} may vary across multiple generations with the same input.
To ensure robustness, we draw \(N\) samples of \(T(\data{t}_i, \data{c}_i)\) and compute their mean, denoted as \(\bar{T}_N(\data{t}_i, \data{c}_i)\):
\begin{equation}
  \label{eqn:evidence_func_n}
  \bar{T}_N(\data{t}_i, \data{c}_i) = \frac{1}{N}\sum_{r=1}^{N}{T^{(r)}(\data{t}_i, \data{c}_i)}
\end{equation}
where \(T^{(r)}(\data{t}_i, \data{c}_i)\) refers to the \(r\)-th sample.
According to \cref{eqn:ephad_sim}, the decision rule for \acrshort{model} is:
\begin{equation}
y = 
  \begin{cases}
    0, & \text{if}\ \exp{-\frac{1}{M_i}\log p(\data{t}_i)}\exp{\bar{T}_N(\data{t}_i, \data{c}_i)/\beta} \geqslant \epsilon \\
    1, & \text{otherwise}
  \end{cases}
\label{eqn:ad}
\end{equation}
 
\mysection{Experiments}
\label{sec:exp}

This section (i) benchmarks \acrshort{model} against existing baselines, (ii) evaluates the impact of behavioral and language patterns on \acrshort{io} user detection via ablation studies, (iii) compares \acrshort{ephad} against alternative methods for integrating these patterns for \acrshort{io} user detection, and (iv) analyzes the sensitivity of \acrshort{model} to different \acrshort{llms} and temperature values \(\beta\).
We run each experiment on A100 and L40S GPUs three times with different random seeds, and their mean and standard deviation (1-sigma) are reported.
The computational complexity of \acrshort{model} for one user is \(O(M_i)\), because the \acrshort{tpp} component analyzes one sequence \(\data{s}_i\) in \(O(M_i)\), while the \acrshort{llm} component processes one sequence in \(O(1)\)\footnote{However, \acrshort{llms} can be the speed bottleneck of \acrshort{model} because they are usually much slower than \acrshort{tpps}. Hence, it is possible that \acrshort{model} takes a constant time to process a sequence, even though the computational complexity is \(O(M_i)\).}.

\acrshort{model} is supposed to be training-free because all its components are either frozen or deterministic computations.
However, to the best of our knowledge, large \acrshort{tpp} models pretrained on large event sequence data do not exist, so we must train a \acrshort{tpp} model on \(\data{T}\) beforehand.
\acrshort{model} works with any existing \acrshort{tpp} model that
provides \(p(\data{t}_i)\).
According to \citep{linExploringGenerativeNeural2022, shchurDetectingAnomalousEvent2021}, the performance gap between existing state-of-the-art \acrshort{tpp} models is small.
Without loss of generality, we choose the \textit{\acrfull{sahp}}\citep{zhangSelfAttentiveHawkesProcess2020} because of its relatively simple design and good performance.
The loss function for training \acrshort{sahp} on \(\data{T}\) is \cref{eqn:nll_of_tpp}.

\acrshort{model} works with all existing \acrshort{llms}.
Recently, we have seen the rise of reasoning models\citep{keSurveyFrontiersLLM2025}.
By enabling the \acrshort{llm} to think during test time, reasoning models consistently improve performance across various tasks, especially for solving mathematical problems and difficult logic problems\citep{openaiOpenAIO1System2024}.
In this paper, we use an open-weight reasoning model \acrshort{gpt} (reasoning model, 117B parameters with 5.1B active parameters)\citep{openaiGptoss120bGptoss20bModel2025} with a medium reasoning effort.
In \cref{sec:exp4}, we report \acrshort{model}'s performance with \acrshort{llama} (non-reasoning model, 70B parameters), \acrshort{qwen} (reasoning model, 80B parameters with 3B active parameters)\citep{qwen3technicalreport}, \acrshort{glm} (reasoning model, 106B parameters with 12B active parameters)\citep{team_glm-45_2025}, and \acrshort{mistral} (non-reasoning model, 24B parameters)\footnote{https://huggingface.co/mistralai/Mistral-Small-3.2-24B-Instruct-2506}.
As for the similarity function in \cref{eqn:evidence_func}, we use the bge-m3-v2 reranker\citep{chenM3EmbeddingMultiLingualityMultiFunctionality2024}.
The \(N\) in \cref{eqn:evidence_func_n} is 5.

\acrshort{model} has one hyperparameter: temperature \(\beta\).
According to Patra et al.\citep{patraEvidenceBasedPostHocAdjustment2025}, we set \(\beta=0.5\) during the experiments.
In \cref{sec:exp4}, we investigate how temperature affects the detection performance of \acrshort{model}.

\partopic{Baseline Models:}
We compare \acrshort{model} with three baselines.
Although not directly aimed at \acrshort{io} user detection, two studies concerning \acrshort{io} users verify coordination among users in the same \acrshort{io}\citep{kongIntervalcensoredTransformerHawkes2023, nwalaLanguageFrameworkModeling2023}. These studies motivate the first two baselines.
\begin{itemize}
    \item \textbf{\acrfull{kong}}\citep{kongIntervalcensoredTransformerHawkes2023} represents each user via a user embedding, generated by averaging the embeddings of all events for that user extracted from a model trained on $\mathcal{T}$. Then, these user-level representations are fed into a clustering model to partition the users into two distinct groups. Users in the smaller group are \acrshort{io} users.
    \item \textbf{\acrfull{nwala}}\citep{nwalaLanguageFrameworkModeling2023} suggests describing user behaviors as strings of symbols using formal languages specified by rules. Next, these sequences are converted into user embeddings using a TF-IDF model. Users whose embeddings differ from the majority are considered \acrshort{io} users.
    \item \textbf{\acrfull{luceri}}\citep{luceriLeveragingLargeLanguage2024} uses \acrshort{llms} to detect in a zero-shot setting whether a user is an \acrshort{io} user based on behavioral and language patterns. This method first checks whether a post \(x_i\) belongs to an \acrshort{io}. Then users whose posts are mostly \acrshort{io} posts are labelled as \acrshort{io} users.
\end{itemize}

\partopic{Datasets:}
We use the \acrshort{io} user dataset collected by Seckin et al.\citep{seckinLabeledDatasetsResearch2025a}.
This dataset contains various identified \acrshort{ios} in 26 labelled datasets across 15 countries.
We evaluate \acrshort{model} and baselines on Egypt, China\_1, Iran\_1, Russia\_1, and UAE.
These datasets have the complete activity timelines of all \acrshort{io} users, but control users are limited to their last 100 posts on days when they are involved in \acrshort{ios}.
To mitigate this bias, we apply the same strategy to \acrshort{io} users by limiting the number of posts from \acrshort{io} users to 100 per day.
\enlargethispage{\baselineskip}
The data statistics of curated datasets are available in \cref{tab:data_statistic}.
We assign 80\% of data to the training set, 10\% to the validation set, and 10\% to the test set.
Please note that although these datasets include labels indicating which users are \acrshort{io} users, they are used solely for performance evaluations. \acrshort{model} and baselines cannot see the labels.

\begin{table}[!tb]
    \caption{Statistics of the curated \acrshort{io} datasets}
    \centering
    \begin{tabular}{lrrrr}
        \toprule
        & \multicolumn{1}{c}{\makecell[c]{Number of\\\acrshort{io} users}} & \multicolumn{1}{c}{\makecell[c]{Number of\\control users}} & \multicolumn{1}{c}{\makecell[c]{Number of\\\acrshort{io} posts}} & \multicolumn{1}{c}{\makecell[c]{Number of\\control posts}} \\
        \midrule
        Egypt     &  219  &  242 & 66,577	 & 14,204 \\
        China\_1  &  537  & 28,445 & 169,287 & 1,718,924 \\
        Iran\_1   &  543  &  3,291 & 247,386 & 192,674 \\
        Russia\_1 &  2,830 &  20,961 & 1,335,064 & 1,595,514 \\
        UAE       &  3,337  &  6,635 & 1,244,984 & 366,873 \\
        \midrule
        Total & 7,466 & 59,574& 3,063,298& 3,888,189 \\
        \bottomrule
    \end{tabular}
    \label{tab:data_statistic}
\end{table}
 
\partopic{Evaluation Metrics:}
We evaluate \acrshort{model} and baselines using precision, recall, F1-score, \acrfull{auc}, and \acrfull{ap}\footnote{\acrshort{ap} is often referred to as Average Precision (AP) in machine learning toolkits. We prefer the term \acrshort{ap} as it more accurately describes the geometric calculation of the metric.}.
We only report \acrshort{ap} for experiments in \cref{sec:exp1}, \cref{sec:exp2}, and \cref{sec:exp4} to simplify the comparison of results.
We add \acrshort{ap} because Davis et al.\citep{davisRelationshipPrecisionRecallROC2006} suggest \acrshort{ap} as an alternative to \acrshort{auc} for tasks with a largely imbalanced class distribution, where \acrshort{auc} results can be overly optimistic.
The \acrshort{io} dataset, as shown in \cref{tab:data_statistic}, is highly imbalanced, with few \acrshort{io} users and many control users, which justifies the use of \acrshort{ap}.

Computing precision, recall, and F1-score requires a threshold \(\epsilon\) to decide the label \(y\).
In this work, the threshold is adjusted by maximizing the F1 on the validation set.
In practice, the scores of \acrshort{io} and control users are often modeled as two class-conditional normal distributions.
The threshold is the score that separates them, that is, most sequences on one side belong to \acrshort{io} users and most sequences on the other side belong to control users.

\mysubsection[Comparing Model against baselines]{Comparing \acrshort{model} with baselines}
\label{sec:exp3}

This section compares \acrshort{model} with three baselines, \acrshort{kong}, \acrshort{nwala}, and \acrshort{luceri} on five \acrshort{io} datasets.
The metrics used are precision, recall, F1-score, \acrshort{auc} and \acrshort{ap}.
The results are presented in \crefrange{tab:io_precision}{tab:io_auprc}.

\begin{table}[!ht]
    \caption{The precision of TENSOR and baselines on five real-world IO datasets (higher is better).}
    \centering
    \begin{tabular}{lcccc}
        \toprule
        & TENSOR & Clustering & BLOC & LLM \\
        \midrule
        Egypt     & \meanstd{0.6138}{0.0055} & \textbf{\meanstd{0.8056}{0.0600}} & \meanstd{0.4761}{0.0000} & \meanstd{0.6882}{0.0316} \\
        China\_1  & \textbf{\meanstd{0.8361}{0.0183}} & \meanstd{0.2649}{0.3495} & \meanstd{0.0321}{0.0000} & \meanstd{0.1190}{0.0018} \\
        Iran\_1   & \meanstd{0.8332}{0.0146} & \textbf{\meanstd{0.9198}{0.0443}} & \meanstd{0.1440}{0.0000} & \meanstd{0.2928}{0.0046} \\
        Russia\_1 & \textbf{\meanstd{0.6644}{0.0290}} & \meanstd{0.1218}{0.0049} & \meanstd{0.1127}{0.0000} & \meanstd{0.2106}{0.0075} \\
        UAE       & \textbf{\meanstd{0.7864}{0.0079}} & \meanstd{0.5824}{0.1452} & \meanstd{0.3357}{0.0000} & \meanstd{0.4556}{0.0048} \\
        \midrule
        Average   & \textbf{0.7468} & 0.5389 & 0.2201 & 0.3532 \\
        \bottomrule
    \end{tabular}
    \label{tab:io_precision}
\end{table}

\begin{table}[!ht]
    \caption{The recall of TENSOR and baselines on five real-world IO datasets (higher is better).}
    \centering
    \begin{tabular}{lcccc}
        \toprule
        & TENSOR & Clustering & BLOC & LLM \\
        \midrule
        Egypt     & \textbf{\meanstd{0.9394}{0.0214}} & \meanstd{0.6818}{0.0371} & \meanstd{0.9091}{0.0000} & \meanstd{0.8636}{0.0000} \\
        China\_1  & \meanstd{0.4691}{0.0175} & \textbf{\meanstd{0.8889}{0.0944}} & \meanstd{0.0926}{0.0000} & \meanstd{0.6605}{0.0087} \\
        Iran\_1   & \meanstd{0.5152}{0.0171} & \meanstd{0.5697}{0.1465} & \textbf{\meanstd{1.0000}{0.0000}} & \meanstd{0.6848}{0.0086} \\
        Russia\_1 & \meanstd{0.6419}{0.0192} & \textbf{\meanstd{0.9435}{0.0725}} & \meanstd{0.8657}{0.0000} & \meanstd{0.2874}{0.0159} \\
        UAE       & \meanstd{0.7784}{0.0098} & \meanstd{0.6766}{0.1045} & \textbf{\meanstd{1.0000}{0.0000}} & \meanstd{0.3832}{0.0000} \\
        \midrule
        Average   & 0.6688 & 0.7521 & \textbf{0.7735} & 0.5759 \\
        \bottomrule
    \end{tabular}
    \label{tab:io_recall}
\end{table}

\begin{table}[!ht]
    \caption{The F1 score of TENSOR and baselines on five real-world IO datasets (higher is better).}
    \centering
    \begin{tabular}{lcccc}
        \toprule
        & TENSOR & Clustering & BLOC & LLM \\
        \midrule
        Egypt     & \meanstd{0.7424}{0.0107} & \meanstd{0.7381}{0.0442} & \meanstd{0.6250}{0.0000} & \textbf{\meanstd{0.7656}{0.0194}} \\
        China\_1  & \textbf{\meanstd{0.6006}{0.0112}} & \meanstd{0.2764}{0.3415} & \meanstd{0.0476}{0.0000} & \meanstd{0.2017}{0.0026} \\
        Iran\_1   & \meanstd{0.6366}{0.0165} & \textbf{\meanstd{0.6895}{0.1178}} & \meanstd{0.2517}{0.0000} & \meanstd{0.4102}{0.0055} \\
        Russia\_1 & \textbf{\meanstd{0.6521}{0.0044}} & \meanstd{0.2154}{0.0055} & \meanstd{0.1995}{0.0000} & \meanstd{0.2430}{0.0103} \\
        UAE       & \textbf{\meanstd{0.7823}{0.0064}} & \meanstd{0.6247}{0.1282} & \meanstd{0.5026}{0.0000} & \meanstd{0.7656}{0.0194} \\
        \midrule
        Average   & \textbf{0.6828} & 0.5088 & 0.3253 & 0.4772 \\
        \bottomrule
    \end{tabular}
    \label{tab:io_f1}
\end{table}

\begin{table}[!ht]
    \caption{The AUC of TENSOR and baselines on five real-world IO datasets (higher is better).}
    \centering
    \begin{tabular}{lcccc}
        \toprule
        & TENSOR & Clustering & BLOC & LLM \\
        \midrule
        Egypt     & \textbf{\meanstd{0.7806}{0.0073}} & \meanstd{0.7570}{0.0395} & \meanstd{0.3636}{0.0000} & \meanstd{0.7558}{0.0306} \\
        China\_1  & \textbf{\meanstd{0.9338}{0.0019}} & \meanstd{0.3645}{0.4207} & \meanstd{0.5872}{0.0000} & \meanstd{0.9057}{0.0064} \\
        Iran\_1   & \textbf{\meanstd{0.8558}{0.0072}} & \meanstd{0.6676}{0.0996} & \meanstd{0.4905}{0.0000} & \meanstd{0.7698}{0.0071} \\
        Russia\_1 & \textbf{\meanstd{0.9213}{0.0026}} & \meanstd{0.3017}{0.0967} & \meanstd{0.3946}{0.0000} & \meanstd{0.7019}{0.0024} \\
        UAE       & \textbf{\meanstd{0.9263}{0.0005}} & \meanstd{0.7403}{0.1254} & \meanstd{0.3426}{0.0000} & \meanstd{0.7558}{0.0306} \\
        \midrule
        Average   & \textbf{0.8836} & 0.5662 & 0.4357 & 0.7778 \\
        \bottomrule
    \end{tabular}
    \label{tab:io_auc}
\end{table}

\begin{table}[!ht]
    \caption{The AUPRC of TENSOR and baselines on five real-world IO datasets (higher is better).}
    \centering
    \begin{tabular}{lcccc}
        \toprule
        & TENSOR & Clustering & BLOC & LLM \\
        \midrule
        Egypt     & \meanstd{0.7690}{0.0087} & \textbf{\meanstd{0.8232}{0.0160}} & \meanstd{0.3839}{0.0000} & \meanstd{0.6859}{0.0284} \\
        China\_1  & \textbf{\meanstd{0.5654}{0.0069}} & \meanstd{0.2712}{0.3694} & \meanstd{0.0288}{0.0000} & \meanstd{0.1670}{0.0053} \\
        Iran\_1   & \textbf{\meanstd{0.6910}{0.0336}} & \meanstd{0.6711}{0.0948} & \meanstd{0.1352}{0.0000} & \meanstd{0.3075}{0.0071} \\
        Russia\_1 & \textbf{\meanstd{0.6952}{0.0133}} & \meanstd{0.0812}{0.0103} & \meanstd{0.0941}{0.0000} & \meanstd{0.1848}{0.0006} \\
        UAE       & \textbf{\meanstd{0.8562}{0.0003}} & \meanstd{0.6783}{0.1474} & \meanstd{0.2475}{0.0000} & \meanstd{0.6859}{0.0284} \\
        \midrule
        Average   & \textbf{0.7154} & 0.5050 & 0.1779 & 0.4062 \\
        \bottomrule
    \end{tabular}
    \label{tab:io_auprc}
\end{table}
 
The results show that \acrshort{model} is the overall best approach for \acrshort{io} user detection, as measured by \acrshort{auc} and \acrshort{ap}.
One outlier is Egypt, where \acrshort{kong} performs better on \acrshort{ap}. 
The reason is the temperature \(\beta\).
Later results in \cref{sec:exp4_2} show that \(\beta=0.5\) from \cite{patraEvidenceBasedPostHocAdjustment2025} is not the best choice.
By lowering the temperature, the overall performance of \acrshort{model} across all datasets significantly improves and can outperform \acrshort{kong} on all datasets.
We do not discuss how to optimize \(\beta\) without any labels in this paper.

\begin{figure}[!tb]
    \centering
    \includegraphics[width=0.9\textwidth]{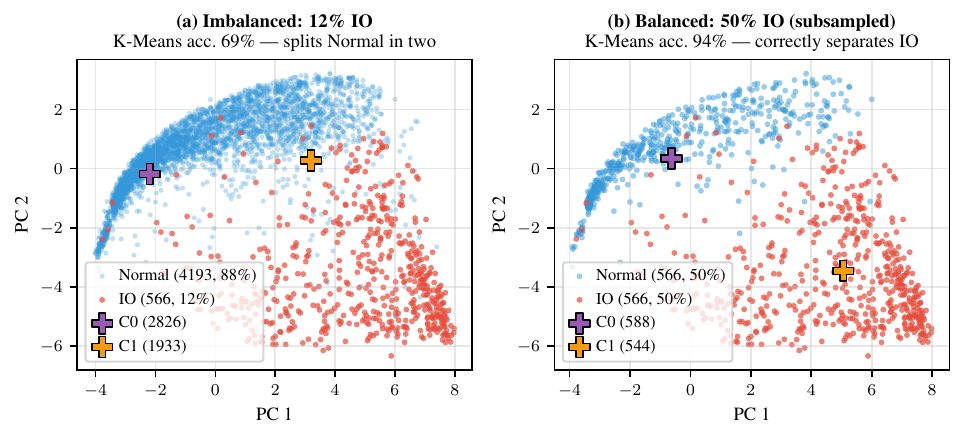}
    \caption{Impact of data imbalance on \acrshort{kong}'s performance. Data points represent control users (C0) and \acrshort{io} users (C1). Results indicate that significant data imbalance shifts cluster centroids with some random seeds, leading to performance inconsistency.}
    \label{fig:kmean}
\end{figure}

We also observe that the results of \acrshort{kong} have a large standard deviation across multiple datasets.
The reason is the data imbalance with many control users and few \acrshort{io} users.
As shown in \cref{fig:kmean}, \acrshort{kong} on imbalanced data is sensitive to random seeds, leading to the misclassification of control users as \acrshort{io} users.
This instability results in inconsistent performance, particularly on highly imbalanced datasets such as China\_1 and Russia\_1.

\subsection[The impact of behavioral and language patterns on IO user detection]{The impact of behavioral and language patterns on \acrshort{io} user detection}
\label{sec:exp1}

If we only use the behavioral or language patterns of one user for \acrshort{io} user detection, we may get suboptimal results.
To demonstrate this, we compare \acrshort{model} with the following ablation baselines: (i) \acrshort{model} without the evidence function \(\bar{T}_N(\data{t}_i, \data{c}_i)\), which is described in \cref{eqn:ad_mtpp}, (ii) \acrshort{model} without \acrshort{tpp}, detecting \acrshort{io} users by \(\bar{T}_N(\data{t}_i, \data{c}_i)\), (iii) \acrshort{model} but \(\bar{T}_N(\data{t}_i, \data{c}_i)\) returns random samples from the uniform distribution \(\mathcal{U}(0, 1)\), and (iv) the Random model, which assigns scores sampled from \(\mathcal{U}(0, 1)\) to users. We include Random because \acrshort{ap} is sensitive to the ratio of control and \acrshort{io} users, whereas the \acrshort{auc} remains at 0.5.
The results are reported in \cref{tab:exp_results_1}.

\begin{table}[!ht]
    \caption{The \acrshort{ap} of \acrshort{model} and ablation baselines on five real-world \acrshort{io} datasets (temperature \(\beta=0.5\), \acrshort{llm} is \acrshort{gpt}, higher is better).}
    \centering
    \resizebox{\textwidth}{!}{\begin{tabular}{lccccc}
        \toprule
        & \acrshort{model} & \makecell{\acrshort{model}\\without \acrshort{llm}} & \makecell{\acrshort{model}\\without \acrshort{tpp}} & \makecell{\acrshort{model}\\with random \(\bar{T}_N(\data{t}_i, \data{c}_i)\)} & Random \\
        \midrule
        Egypt     &  \textbf{\meanstd{0.7690}{0.0087}} & \meanstd{0.6916}{0.0030} & \meanstd{0.5839}{0.0193} & \meanstd{0.6770}{0.0507} & \meanstd{0.5083}{0.0720} \\
        China\_1  &  \textbf{\meanstd{0.5654}{0.0069}} & \meanstd{0.4740}{0.0046} & \meanstd{0.0669}{0.0044} & \meanstd{0.4721}{0.0120} & \meanstd{0.0212}{0.0047} \\
        Iran\_1   &  \textbf{\meanstd{0.6910}{0.0336}} & \meanstd{0.6438}{0.0252} & \meanstd{0.1989}{0.0180} & \meanstd{0.6085}{0.0191} & \meanstd{0.1558}{0.0220} \\
        Russia\_1 &  \textbf{\meanstd{0.6953}{0.0133}} & \meanstd{0.6425}{0.0073} & \meanstd{0.1707}{0.0009} & \meanstd{0.6302}{0.0122} & \meanstd{0.1216}{0.0073} \\
        UAE       &  \textbf{\meanstd{0.8562}{0.0003}} & \meanstd{0.8009}{0.0002} & \meanstd{0.4080}{0.0061} & \meanstd{0.7619}{0.0060} & \meanstd{0.3390}{0.0152} \\
        \midrule
        Average   &  \textbf{0.7154} & 0.6506 & 0.2857 & 0.6299 & 0.2292 \\
        \bottomrule
    \end{tabular}}
    \label{tab:exp_results_1}
\end{table}
 
The ablation results demonstrate that jointly considering behavioral and language patterns for \acrshort{io} user detection is better than considering only one, as the \acrshort{ap} of \acrshort{model} is significantly better than \acrshort{model} without \acrshort{llm} or \acrshort{tpp}.
The experiment results also show that a standalone \acrshort{llm} performs poorly at zero-shot \acrshort{io} user detection.
Another observation is that \acrshort{model} outperforms \acrshort{model} with random \(\bar{T}_N(\data{t}_i, \data{c}_i)\), whose performance is even worse than \acrshort{model} without \acrshort{llm}.
This demonstrates that \acrshort{llm} still provides useful information from \(\data{c}_i\) and \(\data{t}_i\) for \acrshort{io} user detection despite its poor performance in this task.

\mysubsection[The benefit of T(t, c)]{The benefit of \acrshort{llm}-backed \(\bar{T}_N(\data{t}_i, \data{c}_i)\)}
\label{sec:exp2}

\acrshort{ephad} enables \acrshort{model} to jointly consider behavioral and language patterns for \acrshort{io} user detection, but is it the overall best way in terms of performance?
In this section, we investigate several intuitive and existing alternatives to train a \acrshort{io} user detector based on behavioral and language patterns.
Specifically, they are: (i) behavior- and language-aware \acrshort{tpp} model. The input of \acrshort{tpp} is \(\data{t}_i\) and \(\data{c}_i\). The training loss is \(-\log p(\data{t}_i, \data{c}_i)\). Then we use \cref{eqn:ad_mtpp} for \acrshort{io} user detection, (ii) \acrshort{nwala}\citep{nwalaLanguageFrameworkModeling2023}, (iii) \acrshort{luceri}\citep{luceriLeveragingLargeLanguage2024}, and (iv) the Random model.
The results are reported in \cref{tab:exp_results_2}.

\begin{table}[!tb]
    \caption{The \acrshort{ap} of \acrshort{model} and other behavior- and language-aware baselines (temperature \(\beta=0.5\), \acrshort{llm} is \acrshort{gpt}, higher is better).}
    \centering
    \resizebox{\textwidth}{!}{\begin{tabular}{lccccc}
        \toprule
        & \acrshort{model} & \makecell[c]{behavior- and\\language-aware \acrshort{tpp}} & \acrshort{nwala}\citep{nwalaLanguageFrameworkModeling2023} & \acrshort{luceri}\citep{luceriLeveragingLargeLanguage2024} & Random \\
        \midrule
        Egypt     &  \textbf{\meanstd{0.7690}{0.0087}} & \meanstd{0.6949}{0.0017} & \meanstd{0.3838}{0.0160} & \meanstd{0.6859}{0.0284} & \meanstd{0.5083}{0.0720} \\
        China\_1  &  \textbf{\meanstd{0.5654}{0.0069}} & \meanstd{0.4674}{0.0042} & \meanstd{0.0288}{0.0000} & \meanstd{0.1670}{0.0053} & \meanstd{0.0212}{0.0047} \\
        Iran\_1   &  \textbf{\meanstd{0.6910}{0.0336}} & \meanstd{0.6290}{0.0052} & \meanstd{0.1352}{0.0000} & \meanstd{0.3075}{0.0071} & \meanstd{0.1558}{0.0220} \\
        Russia\_1 &  \textbf{\meanstd{0.6953}{0.0133}} & \meanstd{0.6242}{0.0145} & \meanstd{0.0940}{0.0000} & \meanstd{0.1848}{0.0006} & \meanstd{0.1216}{0.0073} \\
        UAE       &  \textbf{\meanstd{0.8562}{0.0003}} & \meanstd{0.8011}{0.0012} & \meanstd{0.2475}{0.0000} & \meanstd{0.4187}{0.0020} & \meanstd{0.3390}{0.0152} \\
        \midrule
        Average   &  \textbf{0.7154} & 0.6433 & 0.1779 & 0.3528 & 0.2292 \\
        \bottomrule
    \end{tabular}}
    \label{tab:exp_results_2}
\end{table}
 
We see that \acrshort{model} outperforms other alternatives by a significant margin on all datasets.
A detailed analysis shows that \acrshort{nwala} performs worse than the Random baseline.
A potential reason is that \acrshort{nwala} requires more detailed information about posts, such as who the post is sent to and whether it contains a picture, which is missing from the dataset.
This results in most generated \acrshort{nwala} sequences containing only one symbol referring to a post, which significantly harms their classification capability.
\acrshort{luceri} and behavior- and language-aware \acrshort{tpp} outperform Random but are outperformed by \acrshort{model}.
On closer inspection, we find that the results of the behavior- and language-aware \acrshort{tpp} are basically the same as those of \acrshort{tpp}.
These results demonstrate that adding language data to the \acrshort{tpp} model does not improve its \acrshort{io} detection capability.

\mysubsection[Sensitivity]{Sensitivity to different \acrshort{llms} and \(\beta\)}
\label{sec:exp4}

All reported results of \acrshort{model} in previous sections are based on temperature \(\beta=0.5\) inherited from the \acrshort{ephad} paper and \acrshort{gpt}.
However, \acrshort{model} works with all existing \acrshort{llms} and temperatures \(\beta > 0\).
In this section, we evaluate the stability of \acrshort{model} under different \acrshort{llms} and temperatures.

\partopic{Sensitivity to \acrshort{llms}:}
\label{sec:exp4_1}

\begin{table}[!tb]
    \caption{The \acrshort{ap} of \acrshort{model} with different \acrshort{llms} in \(\bar{T}_N(\data{t}_i, \data{c}_i)\) (temperature \(\beta=0.5\), higher is better).}
    \centering
    \resizebox{\textwidth}{!}{\begin{tabular}{lcccccc}
        \toprule
        & \makecell{\acrshort{model} w/\\\acrshort{gpt}} & \makecell{\acrshort{model} w/\\ \acrshort{llama}} & \makecell{\acrshort{model} w/\\  \acrshort{qwen}}  & \makecell{\acrshort{model} w/\\  \acrshort{glm}} & \makecell{\acrshort{model} w/\\  \acrshort{mistral}}  & \makecell{\acrshort{model}\\without \acrshort{llm}} \\
        \midrule
        Egypt     &  \textbf{\meanstd{0.7690}{0.0087}} & \meanstd{0.7079}{0.0019} & \meanstd{0.7373}{0.0084} & \meanstd{0.7467}{0.0017} & \meanstd{0.6806}{0.0057} & \meanstd{0.6916}{0.0030} \\
        China\_1  &  \textbf{\meanstd{0.5654}{0.0069}} & \meanstd{0.5224}{0.0062} & \meanstd{0.5251}{0.0030} & \meanstd{0.5609}{0.0035} & \meanstd{0.5085}{0.0044} & \meanstd{0.4740}{0.0046}  \\
        Iran\_1   &  \textbf{\meanstd{0.6910}{0.0336}} & \meanstd{0.6622}{0.0269} & \meanstd{0.6711}{0.0948} & \meanstd{0.6737}{0.0345} & \meanstd{0.6480}{0.0353} & \meanstd{0.6438}{0.0252}  \\
        Russia\_1 &  \textbf{\meanstd{0.6953}{0.0133}} & \meanstd{0.6530}{0.0103} & \meanstd{0.6358}{0.0131} & \meanstd{0.6475}{0.0122} & \meanstd{0.6495}{0.0081} & \meanstd{0.6425}{0.0073} \\
        UAE       &  \textbf{\meanstd{0.8562}{0.0003}} & \meanstd{0.8154}{0.0022} & \meanstd{0.7850}{0.0018} & \meanstd{0.8152}{0.0007} & \meanstd{0.7804}{0.0009} & \meanstd{0.8009}{0.0002} \\
        \midrule
        Average   &  \textbf{0.7154} & 0.6670 & 0.6709 & 0.6888 & 0.6534 & 0.6506 \\
        \bottomrule
    \end{tabular}}
    \label{tab:exp_results_4_llm}
\end{table}
 
\cref{tab:exp_results_4_llm} shows the performance of \acrshort{model} with different \acrshort{llms}.
Besides \acrshort{gpt}, we pick \acrshort{llama}, \acrshort{qwen}, \acrshort{glm}, and \acrshort{mistral}.
We observe that configurations using \acrshort{gpt}, \acrshort{llama}, and \acrshort{glm} consistently outperform \acrshort{model} without \acrshort{llm} across all datasets.
Although \acrshort{qwen} and \acrshort{mistral} underperform on some datasets because of their smaller model or active parameter sizes, their overall average is still better than \acrshort{model} without \acrshort{llm}.
These results show that \acrshort{model} is stable across different \acrshort{llms}.

\partopic{Sensitivity to \(\beta\):}
\label{sec:exp4_2}

\begin{figure}[ht]
    \centering
    \begin{minipage}{0.19\textwidth}
        \centering
        \includegraphics[width=\textwidth]{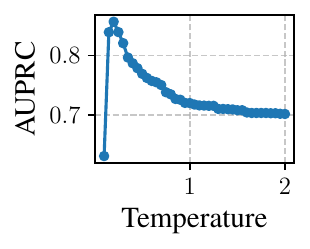}
        Egypt
    \end{minipage}
    \begin{minipage}{0.19\textwidth}
        \centering
        \includegraphics[width=\textwidth]{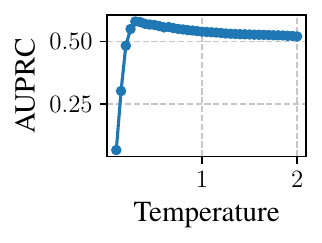}
        China\_1
    \end{minipage}
    \begin{minipage}{0.19\textwidth}
        \centering
        \includegraphics[width=\textwidth]{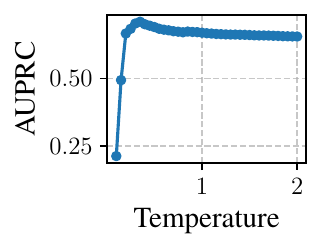}
        Iran\_1
    \end{minipage}
    \begin{minipage}{0.19\textwidth}
        \centering
        \includegraphics[width=\textwidth]{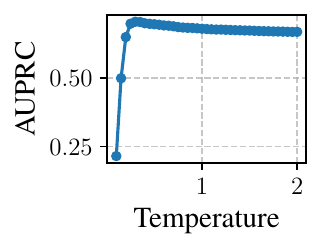}
        Russia\_1
    \end{minipage}
    \begin{minipage}{0.19\textwidth}
        \centering
        \includegraphics[width=\textwidth]{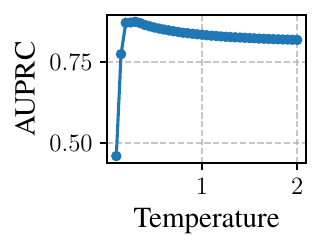}
        UAE
    \end{minipage}

    \caption{\acrshort{ap} of \acrshort{model} on all five \acrshort{io} datasets with different temperatures from 0.05 to 2.}

    \label{fig:temperature}
\end{figure}

\cref{fig:temperature} shows how temperature \(\beta\) affects the performance of \acrshort{model}.
We observe that although the optimal temperature varies across datasets, it is consistently around 0.3 for \acrshort{gpt}.
Generally, increasing the temperature from 0.05 brings a quick performance improvement to a maximum, followed by a gradual decline.
For China\_1, Iran\_1, Russia\_1, and UAE, the performance drop is relatively small, while for Egypt, the \acrshort{ap} drops significantly from over 0.85 to below 0.70.
Table \ref{tab:exp_results_4_best_temp} compares \acrshort{model} at its optimal temperature against the default configuration and other baselines.
\acrshort{model} consistently outperforms other approaches at the optimal temperature.
These results indicate that unsupervised optimization of \(\beta\) is a promising approach to further improve the performance of \acrshort{model}.
We leave this as future work.

 \begin{table}[!ht]
    \caption{The \acrshort{ap} of \acrshort{model} and other behavior- and language-aware baselines (temperature \(\beta=0.5\) for \acrshort{model} results in the second column, \acrshort{llm} is \acrshort{gpt}, higher is better).}
    \centering
    \resizebox{\textwidth}{!}{\begin{tabular}{lcccccc}
        \toprule
        & \makecell[c]{\acrshort{model}\\with best \(\beta\)} & \acrshort{model} & \acrshort{kong}\citep{kongIntervalcensoredTransformerHawkes2023} & \acrshort{nwala}\citep{nwalaLanguageFrameworkModeling2023} & \acrshort{luceri}\citep{luceriLeveragingLargeLanguage2024} & Random \\
        \midrule
        Egypt     &  \textbf{\meanstd{0.8568}{0.0081}}  &  \meanstd{0.7690}{0.0087} & \meanstd{0.8232}{0.0160} & \meanstd{0.3838}{0.0160} & \meanstd{0.6859}{0.0284} & \meanstd{0.5083}{0.0720} \\
        China\_1  &  \textbf{\meanstd{0.5796}{0.0175}}  &  \meanstd{0.5654}{0.0069} & \meanstd{0.2712}{0.3694} & \meanstd{0.0288}{0.0000} & \meanstd{0.1670}{0.0053} & \meanstd{0.0212}{0.0047} \\
        Iran\_1   &  \textbf{\meanstd{0.7108}{0.0271}}  &  \meanstd{0.6910}{0.0336} & \meanstd{0.6711}{0.0948} & \meanstd{0.1352}{0.0000} & \meanstd{0.3075}{0.0071} & \meanstd{0.1558}{0.0220} \\
        Russia\_1 &  \textbf{\meanstd{0.7045}{0.0092}}  &  \meanstd{0.6953}{0.0133} & \meanstd{0.0812}{0.0103} & \meanstd{0.0940}{0.0000} & \meanstd{0.1848}{0.0006} & \meanstd{0.1216}{0.0073} \\
        UAE       &  \textbf{\meanstd{0.8743}{0.0007}}  &  \meanstd{0.8562}{0.0003} & \meanstd{0.6783}{0.1474} & \meanstd{0.2475}{0.0000} & \meanstd{0.4187}{0.0020} & \meanstd{0.3390}{0.0152} \\
        \midrule
        Average   &  \textbf{0.7452} &  0.7154 & 0.5050 & 0.1779 & 0.3528 & 0.2292 \\
        \bottomrule
    \end{tabular}}
    \label{tab:exp_results_4_best_temp}
\end{table}
  
\mysection[Conclusion]{Conclusion}
\label{sec:conclusion}

In this study, we addressed the critical challenge of identifying \acrshort{io} users within social networks. While traditional supervised models struggle with the evolving nature of \acrshort{io} user behaviors, unsupervised models rely on rigid assumptions of coordination, \acrshort{model} offers a more resilient alternative by framing detection as a multimodal anomaly problem.
\acrshort{model} shifts the focus toward the inherent friction between \acrshort{io} campaigns and organic user behaviors. By deploying a \acrshort{tpp}, we successfully captured the behavioral patterns that distinguish coordinated influence from genuine social interaction. A key contribution of this work is the mitigation of ``training set contamination.'' By integrating a novel evidence function that translates \acrshort{llm} responses, which are generated from users' post timelines, into quantitative scores, we effectively adjust the output of \acrshort{tpp} to mitigate the noise caused by \acrshort{io} users embedded in the training data.
Experimental results show that \acrshort{model} outperforms other baselines by a significant margin on five real-world \acrshort{io} datasets. 
\begin{credits}
\subsubsection{\ackname}
This research is supported in part by the Australian Research Council (ARC) Discovery Projects DP200101441 and DP210100743. We appreciate the compute and LLM services provided by RMIT RACE Hub.

\end{credits}
 
\bibliography{reference}

\end{document}